\documentclass[conference]{IEEEtran}

\usepackage{float}
\usepackage{multirow, multicol}
\usepackage{graphicx}
\usepackage{hhline}
\usepackage{amsmath}
\usepackage{array, makecell}%
\usepackage{bm}
\usepackage{amsbsy}
\usepackage{xspace}
\usepackage{xcolor}
\usepackage{newfloat}
\usepackage[export]{adjustbox}
\usepackage{enumitem}
\usepackage[hyphens]{url}
\usepackage{subcaption}
\usepackage{float}
\usepackage{multirow, multicol}
\usepackage{graphicx}
\usepackage{hhline}
\usepackage{amsmath}
\usepackage{array, makecell}%
\usepackage{bm}
\usepackage{amsbsy}
\usepackage{booktabs}
\usepackage{xspace}
\usepackage{xcolor}
\usepackage{newfloat}
\usepackage[export]{adjustbox}
\usepackage{enumitem}
\usepackage{url}
\usepackage{times}
\usepackage{epsfig}
\usepackage{graphicx}
\usepackage{amsmath}
\usepackage{amssymb}
\usepackage{tabularx}
\usepackage[hidelinks]{hyperref}
\usepackage{tikz}

\def\eg{{e.g.,}\@\xspace} 
\def\ie{{i.e.,}\@\xspace}

\newcommand\copyrighttext{%
  \footnotesize \textcopyright 2024 IEEE. Personal use of this material is permitted. Permission from IEEE must be obtained for all other uses, in any current or future media, including reprinting/republishing this material for advertising or promotional purposes, creating new collective works, for resale or redistribution to servers or lists, or reuse of any copyrighted component of this work in other works.
  DOI: \href{https://doi.org/10.1109/IWBF62628.2024.10593873}{10.1109/IWBF62628.2024.10593873}}
\newcommand\copyrightnotice{%
\begin{tikzpicture}[remember picture,overlay]
\node[anchor=south,yshift=10pt] at (current page.south) {\fbox{\parbox{\dimexpr\textwidth-\fboxsep-\fboxrule\relax}{\copyrighttext}}};
\end{tikzpicture}%
}

\begin{document}

\title{Multi-Channel Cross Modal Detection of Synthetic Face Images}

\author{\IEEEauthorblockN{M. Ibsen$^1$, C. Rathgeb$^1$, S. Marcel$^2$, C. Busch$^1$}
\IEEEauthorblockA{1 - Biometrics and Security Research Group, Hochschule Darmstadt, 64295 Darmstadt, Germany \\
mathias.ibsen@h-da.de}
\IEEEauthorblockA{2 - Biometrics Security and Privacy Group, Idiap Research Institute, 1920 Martigny, Switzerland}
}

\maketitle
\copyrightnotice

\IEEEpubidadjcol

\begin{abstract}
Synthetically generated face images have shown to be indistinguishable from real images by humans and as such can lead to a lack of trust in digital content as they can, for instance, be used to spread misinformation. Therefore, the need to develop algorithms for detecting entirely synthetic face images is apparent. Of interest are images generated by state-of-the-art deep learning-based models, as these exhibit a high level of visual realism. Recent works have demonstrated that detecting such synthetic face images under realistic circumstances remains difficult as new and improved generative models are proposed with rapid speed and arbitrary image post-processing can be applied. In this work, we propose a multi-channel architecture for detecting entirely synthetic face images which analyses information both in the frequency and visible spectra using Cross Modal Focal Loss. We compare the proposed architecture with several related architectures trained using Binary Cross Entropy and show in cross-model experiments that the proposed architecture supervised using Cross Modal Focal Loss, in general, achieves most competitive performance.
\end{abstract}

\section{Introduction}
\label{sec:intro}
Artificial intelligence (AI) has seen rapid development in the last few years, especially notably within generative models, which today are capable of synthesizing realistic digital content such as image and text which are almost indistinguishable from real data by humans~\cite{Nightingale-HumanDetectionOfAiSynthesizedImages-PNAS-2022}. Such synthetic data can have malicious use, \eg when used for creating fake news and for performing financial fraud~\cite{Kietzmann-DeepfakesTrickOrTreat-BusinessHorizons-2020}.

\begin{figure}[!t]
\centering
\includegraphics[width=0.70\linewidth]{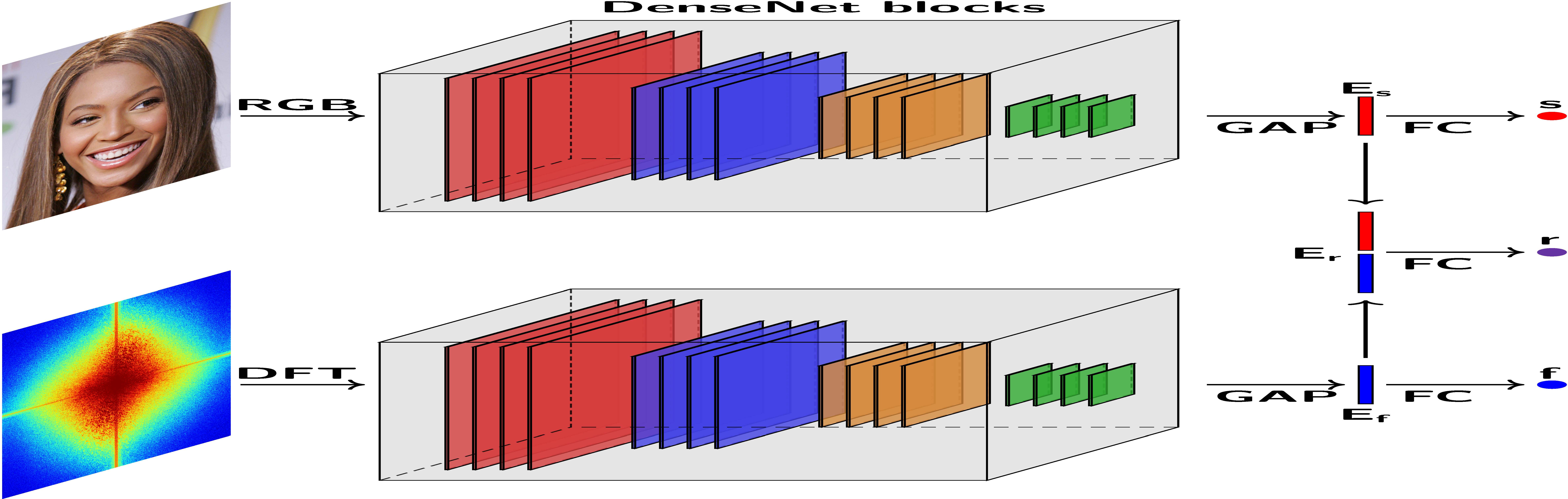}
\caption{Proposed multi-channel architecture for detecting synthetic face images. A RGB face image and its corresponding frequency spectra are fed into separate network channels based on DenseNet. Global Average Pooling (GAP) is applied to the output of each channel to obtain embeddings which are concatenated into a joint representation. For each embedding, a fully connected layer (FC) is added together with the Sigmoid function, which results in three network heads. We propose to supervise the network using Cross Modal Focal Loss.} \vspace{-0.4cm} 
\label{fig:arch_overview}
\end{figure}

With today's mainly free flow of information in the digital realm, it is crucial to be able to distinguish fake digital content from pristine digital content. Of particular interest is the detection of digitally generated images. Visual stimuli have traditionally been associated with reliable information and popularized the phrase \textit{seeing is believing}. However, face images and, in particular, the so-called deepfakes, \ie images and videos manipulated using advanced deep learning-based methods, has meant that this is no longer veracious. For long, a lot of focus was put on discovering images which were manipulated manually using dedicated image manipulation tools~\cite{Rathgeb2019-beaty}. However, the emerging realism and availability of generative models has meant that realistic synthetic media can be generated automatically with high realism and ease~\cite{Joshi-SyntheticDataInHumanAnalysisSurvey-2022}. 

It has been shown that it is possible to detect synthetic images under the assumption that the generative model is known~\cite{Wang-CNNGeneratedImagesAreSurprisinglyEasyToSpotForNow-CVPR-2020}. However, detecting synthetic images across synthesis techniques is more complex, especially if both pristine and fake images undergo realistic post-processing operations. Despite efforts to improve face forgery detection capabilities, recent studies show that generalization is still an ongoing and constant struggle~\cite{Gragnaniello-DetectionOfAIGeneratedFaces-Springer-2022}. Therefore, there is a need to continue to explore ways to improve detection performance and put forth guidelines on how to detect synthetic face images generated with unseen methods. Current studies for detecting synthetic face images usually either learn features trained directly on the RGB images or transform the images to the frequency spectrum whereafter features are extracted. Only a few works have explored the use of both frequency and RGB information for detecting synthetic face images. In this work, we propose a multi-channel architecture that uses both information from the frequency and visible spectra to detect synthetic face images,  as illustrated in figure~\ref{fig:arch_overview}. Moreover, we propose to supervise the training of the network by using the Cross Modal Focal Loss (CMFL)~\cite{George-CMFLForRGBDFaceAntiSpoofing-CVPR-2021} as auxiliary learning for the separate network channels. We compare the proposed architecture to relevant single-channel and multi-channel baselines. The code for the proposed method is made available~\footnote{\url{https://github.com/dasec/Multi-Channel-Cross-Modal-Detection-of-Synthetic-Face-Images}}.

This paper makes the following contributions:

\begin{itemize}
    \item A multi-channel architecture for detecting synthetic face images using the RGB and frequency spectra
    \item The use of CMFL for detecting synthetic face images
    \item An extensive evaluation of the generalizability capability of the proposed architecture, including a comparison to related architectures and loss functions
\end{itemize}

\section{Related Work}
\label{sec:related_work}
\label{sec:synthetic_face_detection}
Methods for detecting synthetic face images usually rely on artefacts or statistical differences in pixel intensities or textures between pristine and fake images. Most works prior to the deep learning-based methods relied on handcrafted features, \eg by exploiting visual artefacts such as missing reflections, lack of geometry, and differences in eye color~\cite{Matern-ExploitingVisualArtefacts-WACVW-2019} or by analyzing differences in landmark localization~\cite{Yang-ExposingGANSynthesizedFacesUsingLandmarks-ACM-2019}.
Nowadays, most approaches are deep learning-based and rely on large image datasets to learn discriminative features that allow distinguishing between pristine and fake images. Early works, such as~\cite{roessler2019faceforensics}, showed that it was possible to detect synthetic face images in the visible spectra using Convolutional Neural Networks (CNNs) when images were generated by the same model. However, they did not explore the performance on images stemming from generative models not seen during training. In~\cite{Wang-CNNGeneratedImagesAreSurprisinglyEasyToSpotForNow-CVPR-2020}, the authors show that by using data augmentation and training on only ProGAN-generated images, it is possible to achieve surprisingly high performance with a mean average precision (AP) higher than 90\% for some training configurations when evaluated on 11 different CNN-based image generators. 

Besides detecting images in the visible spectrum it has also been shown that it is possible to detect synthetic face images in the frequency domain by taking advantage of unnatural firm peaks in the frequency spectrum which, especially, can occur due to the up-sampling performed by the generative model~\cite{Gragnaniello-DetectionOfAIGeneratedFaces-Springer-2022}. In~\cite{Zhang-AutoGAN-WIFS-2019}, the authors propose to detect images in the frequency spectrum instead of the spatial domain after observing that artefacts are manifested in the frequency domain of synthetic images. It has been shown, \eg in~\cite{Frank-FrequencyAnalysisForDeepFakeImageRecognition-ICML-2020}, that such artefacts in the frequency spectrum can be consistent across different generative model architectures. 

While multi-channel methods have already been proposed for detecting synthetic and tampered face images, they are typically proposed for other types of digital manipulations, \eg identity swaps~\cite{Masi-TwoBranchRecurrentNetworkForIsolatingDeepfakesInVideos-ECCV-2020,Hsu-DeepfakeAlgorithmUsingMultiplNoiseModalitiesWithTwoBranchPrediction-APSIPAASC} or the learning scenario is different as they either consider score-level~\cite{Chen-JoiuntSpatialFrequencyDomainNetwork-arxiv-2020, Zhou-TwoStreamTamperedFaceDetection-CVPRW-2017} or feature-level~\cite{Fu-RobustDualChannelGANDetectionBasedOnFilters-CISP-2019} fusion of the branches. These learning scenarios differ from the proposed framework, which solves the task using CMFL for supervision of the individual branches and Binary Cross Entropy (BCE) for the joint architecture head, as further explained in the subsequent section.

\section{Proposed Method}
\label{sec:proposed_approach}
As mentioned, both the spatial and frequency domains have shown to be helpful in detecting synthetic face images. However, while the two domains convey the same underlying data, it represents different views of the data, which can be useful for detection under different circumstances. Therefore, the proposed solution is a multi-channel CNN with separate channels for the RGB and frequency domain. To supervise this network, CMFL is used. 

\subsection{Architecture}
The architecture of the multi-channel network is shown in figure~\ref{fig:arch_overview}. As seen, the network consists of two separate network branches; one branch based on the RGB input image and another branch based on DFT of each colour channel. Each branch consists of the first eight blocks from the DenseNet161 architecture~\cite{Huang-DenseNet-CVPR-2017} initialized using pretrained weights learned on the ImageNet dataset. After each branch, global average pooling is applied to obtain a 384-dimensional embedding layer which are concattenated to form a joint embedding for the entire network. A fully connected layer plus the Sigmoid activation function is added on top of each embedding layer to form the different heads of the architecture which is required by the CMFL loss function as explained below. In this work, we use DenseNet blocks for the model architecture, but this can easily be exchanged for other backbones, \eg ResNet and MobileNet. A comprehensive investigation of the most optimal backbone to use is deliberately left out of this work.

\subsection{Cross Modal Focal Loss}

CMFL is based on Focal Loss~\cite{Lin-FocalLoss-ICCV-2017} and has previously been used for presentation attack detection (PAD)~\cite{George-CMFLForRGBDFaceAntiSpoofing-CVPR-2021}. The motivation to use CMFL for detecting synthetic face images is that in cases where one channel classifies a sample with high confidence, the loss contribution of the other channel can be reduced. The assumption is that this is useful for detecting synthetic face images as sometimes artefacts manifest in the RGB domain, and other times, it is easier to detect synthetic images in the frequency domain. Hence, if it is possible to detect synthetic images using one channel, we do not want the other channel to penalize the model.

To introduce the used loss functions we start by defining cross entropy (CE) for binary classification which can be expressed as: 

\begin{equation}
        CE(p,y) = 
\begin{cases}
     -\log(p) & \text{if } y = 1\\
   -\log(1-p)              & \text{otherwise}
\end{cases}
    \label{eq:bce}
\end{equation}

where $y=1$, in our case, denotes a pristine image and $y=0$ denotes a synthetic image and where p is the probability of the class. The probability of a target class, $p_t$ can be simplied as:

\begin{equation}
        p_t = 
\begin{cases}
     p & \text{if } y = 1\\
     1-p              & \text{otherwise}
\end{cases}
\end{equation}

hence we can now rewrite equation~\ref{eq:bce} as: $CE(p,y) = CE(p_t) = -\log(p_t)$. Focal Loss (FL) can then be expressed as follows: 

\begin{equation}
    FL(p_t) = -\alpha_t(1 - p_t)^\gamma \log(p_t)
\end{equation}

which is similar to the rewritten CE loss, but where $(1 - p_t)^\gamma$ acts as a modulating factor. Here, $\gamma \geq 0$ acts as a tune-able focusing parameter and $\alpha$ is a weighing term added to each class.  The purpose of the modulating term introduced in FL is to focus the learning more on difficult samples.

The idea of CMFL is to extend this to a multi-channel network architecture with two separate input branches. Consider figure~\ref{fig:arch_overview} where spatial (RGB) and frequency (DFT) information is given to two separate network channels, in our case DenseNet blocks. Furthermore let $E_s$ and $E_f$ be the embeddings obtained from computing the forward pass for the spatial and frequency domains, respectively. We can then compute $E_r$ to be a joint representation of $E_s$ and $E_f$ such that we have three output embeddings, one for each channel and a combined embedding. A separate model head is then added to each embedding by adding a single fully connected dense layer followed by a Sigmoid layer after each embedding layer (see figure \ref{fig:arch_overview}). The output probabilities of each of these heads can then be denoted as $s,f,r$ where $s$ and $f$ are the probabilities of each of the branches separately and $r$ is the probability for the joint architecture head. Given this notation, CMFL is defined as given in equation~\ref{eq:cmfl}. 

\begin{equation}
    CMFL_{s_t,f_t} = -\alpha_t(1 - w(s_t, f_t))^\gamma \log(s_t)
    \label{eq:cmfl}
\end{equation}

Where $w(s_t, f_t)$ is dependent on the probabilities given of the separate channels as expressed in equation~\ref{eq:channel_w} and where $\alpha$ and $\gamma$ are the coefficients from FL. 

\begin{equation}
    w(s_t, f_t) = f_t \frac{2 s_t f_t}{s_t + f_t}
    \label{eq:channel_w}
\end{equation}

For the final loss, CMFL is used as auxiliary supervision of the individual branches combined with CE for the joint architecture head. The final loss then becomes:

\begin{equation}
    \mathcal{L} = (1 - \lambda)\mathcal{L}_{CE(r_t)} + \lambda(\mathcal{L}_{CMFL_{s_t,f_t}} + \mathcal{L}_{CMFL_{f_t,s_t}})
    \label{eq:aux_cmfl}
\end{equation}

where $\lambda$ is a changeable coefficient.

\section{Experimental Setup}
\label{sec:experimental_setup}

\subsection{Data}
\begin{figure}[!htbp]
    \centering
    \includegraphics[width=\linewidth]{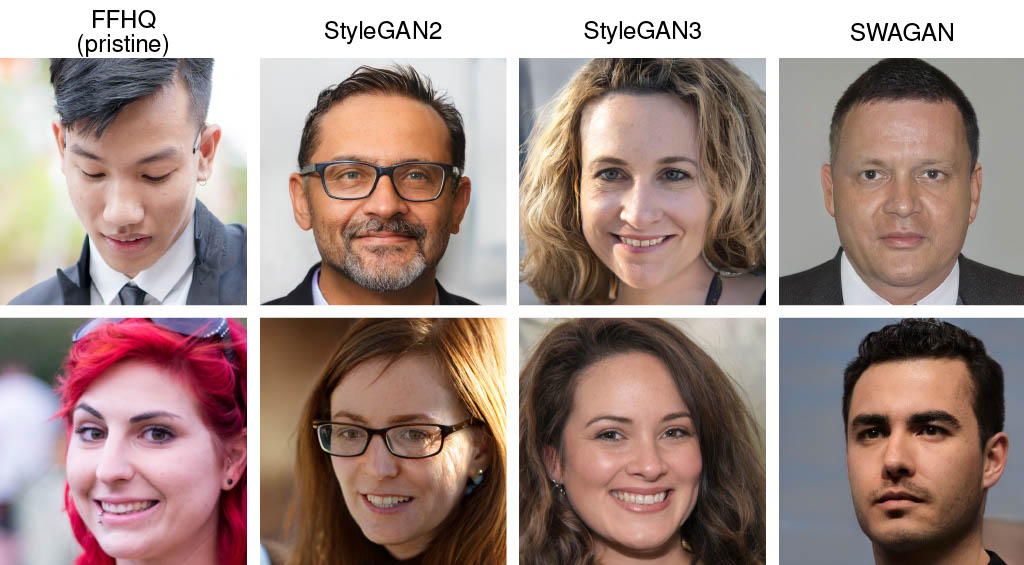}
    \caption{Examples of pristine and synthetic images from the different used generative models.}
    \label{fig:db_examples}
\end{figure}

To evaluate the multi-channel architecture and loss functions, we use pristine images from the Flickr-Faces-HQ dataset (FFHQ)~\cite{ffhq} and synthetic images from StyleGAN2~\cite{Karras-StyleGAN2-CVPR-2020}, StyleGAN3~\cite{Karras-StyleGAN3-NIPS-2021} and SWAGAN~\cite{Gal-Swagan-ACM-2021}. Each of the generative models has been trained on FFHQ; for SWAGAN the model from~\cite{Kabbani-EGAIN-2022-EUVIP} is used. For each model, 50,000 images were generated using seeds 0-49,999 with a truncation factor of 1. Examples of pristine and synthetic data are given in figure~\ref{fig:db_examples}.

\begin{figure}[!tb]
\begin{subfigure}{0.24\linewidth}
    \centering %
  \includegraphics[width=\textwidth]{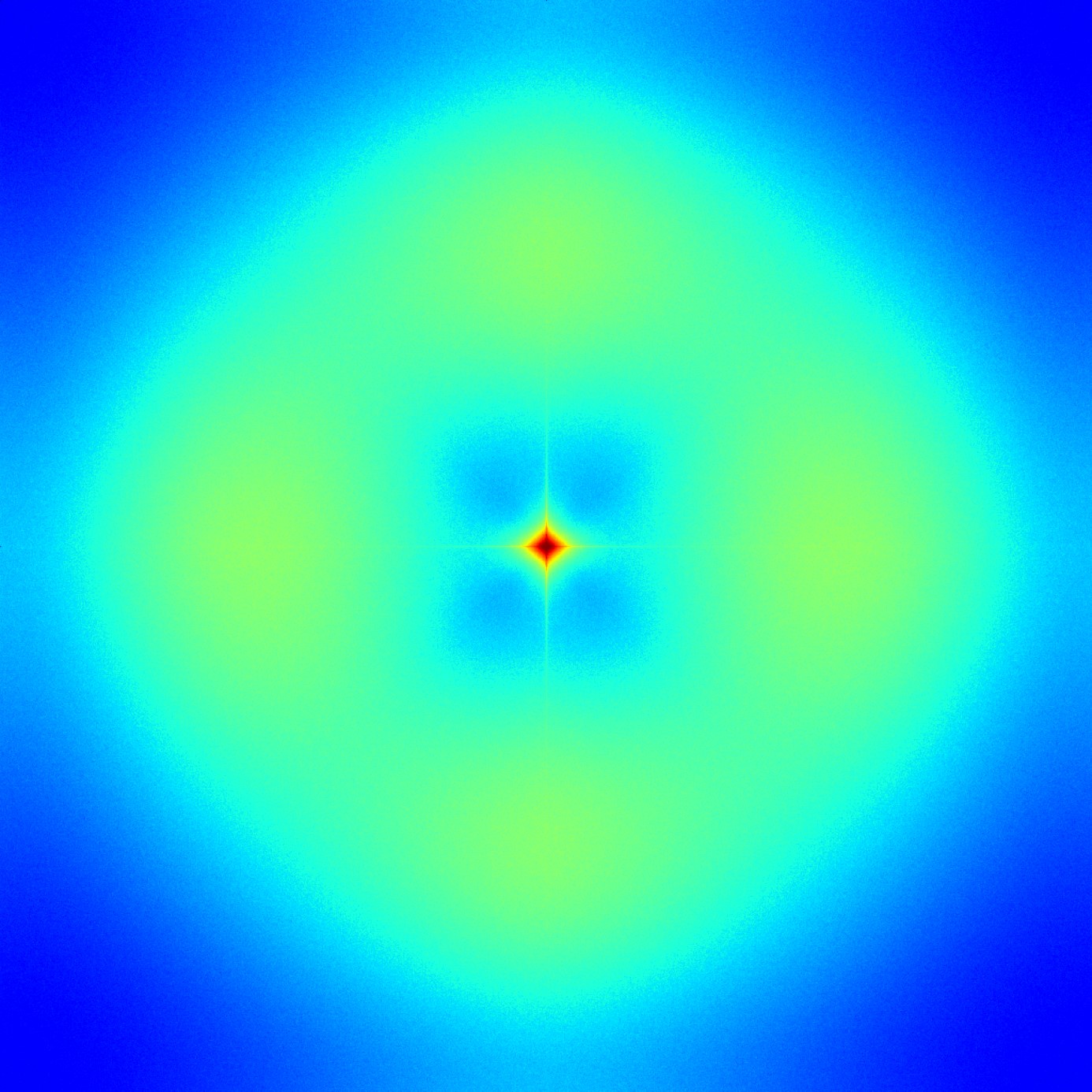}
  \caption{FFHQ}
\end{subfigure}
\begin{subfigure}{0.24\linewidth}
    \centering 
  \includegraphics[width=\textwidth]{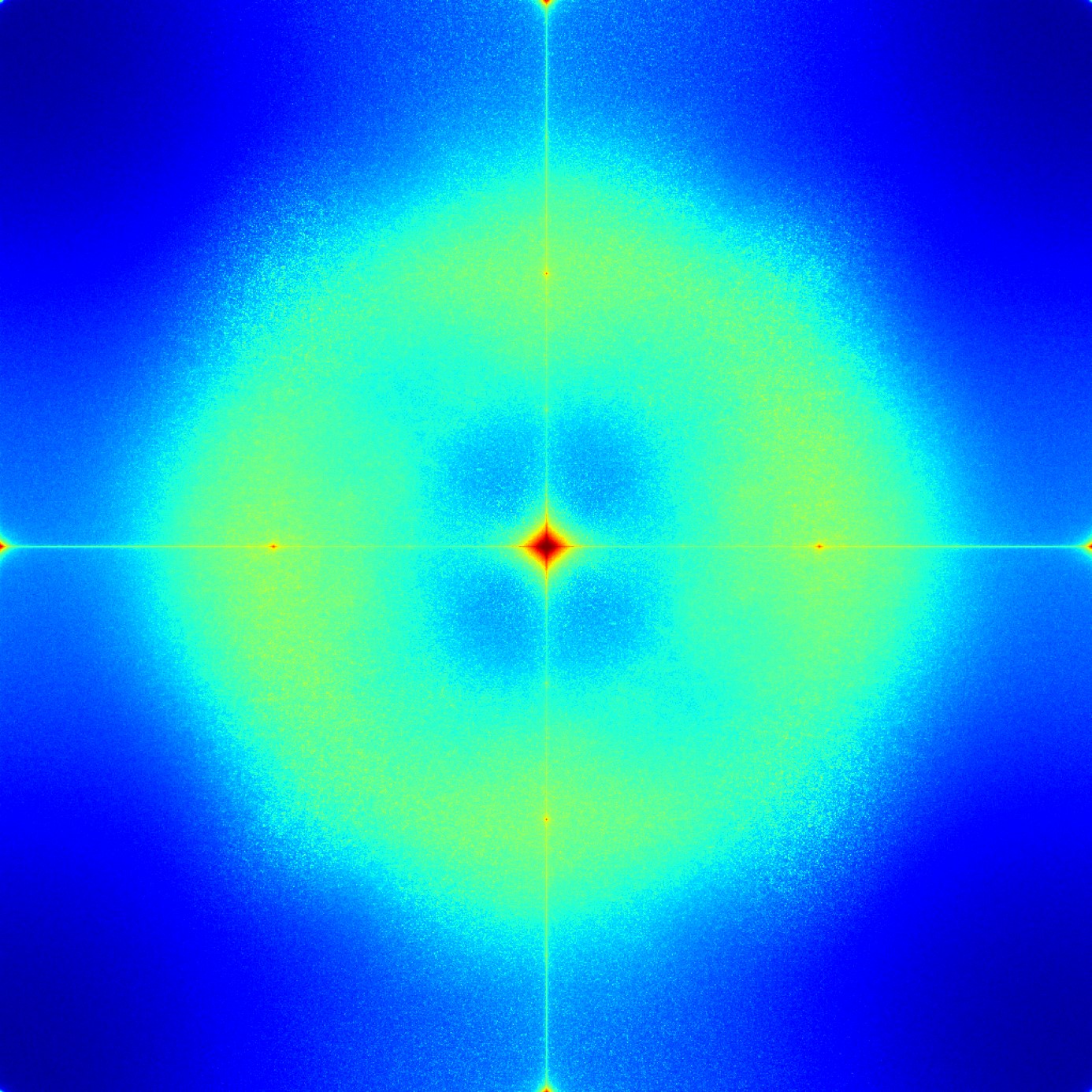}
  \caption{StyleGAN2}
\end{subfigure}
\begin{subfigure}{0.24\linewidth}
    \centering %
  \includegraphics[width=\textwidth]{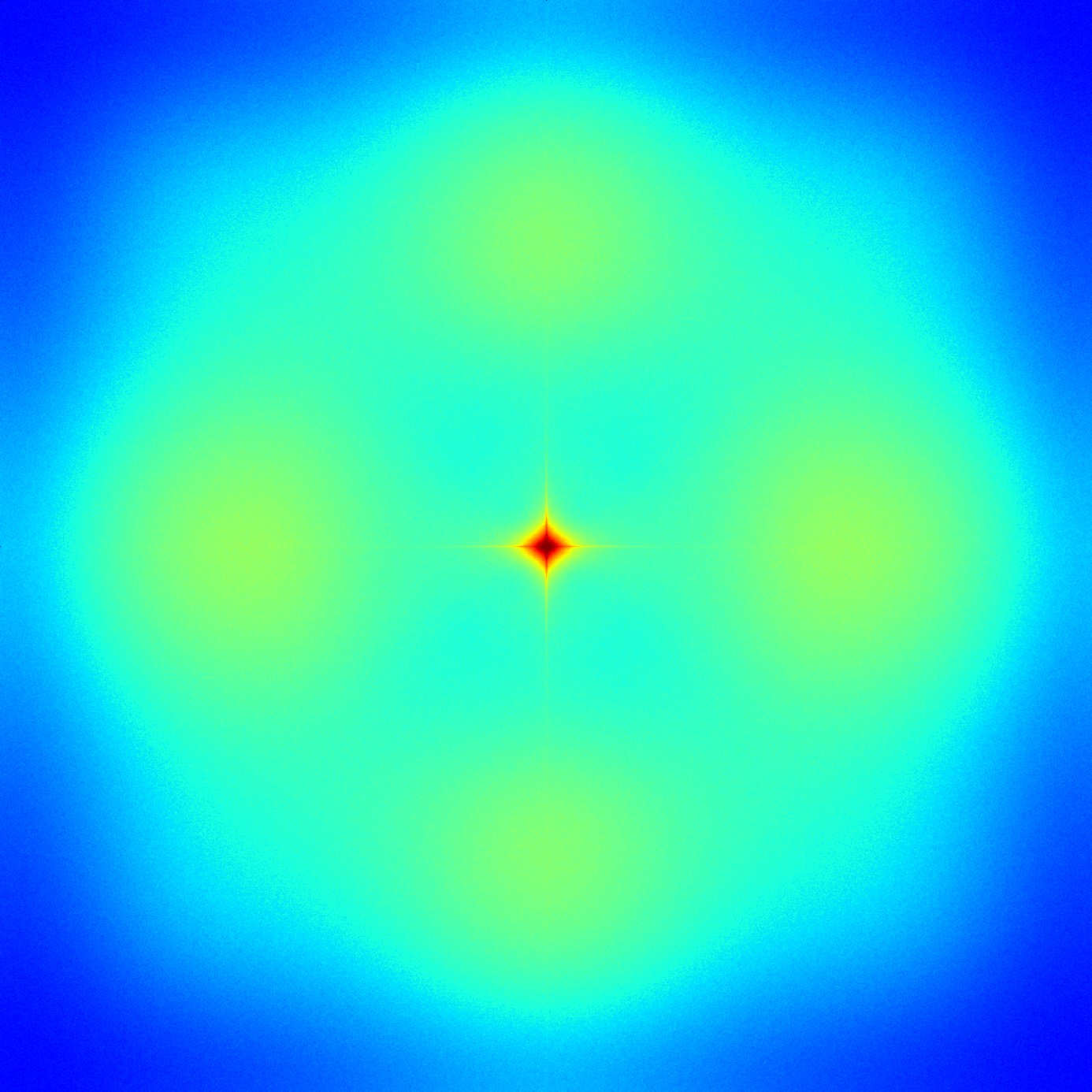}
  \caption{StyleGAN3}
\end{subfigure}
\begin{subfigure}{0.24\linewidth}
    \centering %
  \includegraphics[width=\textwidth]{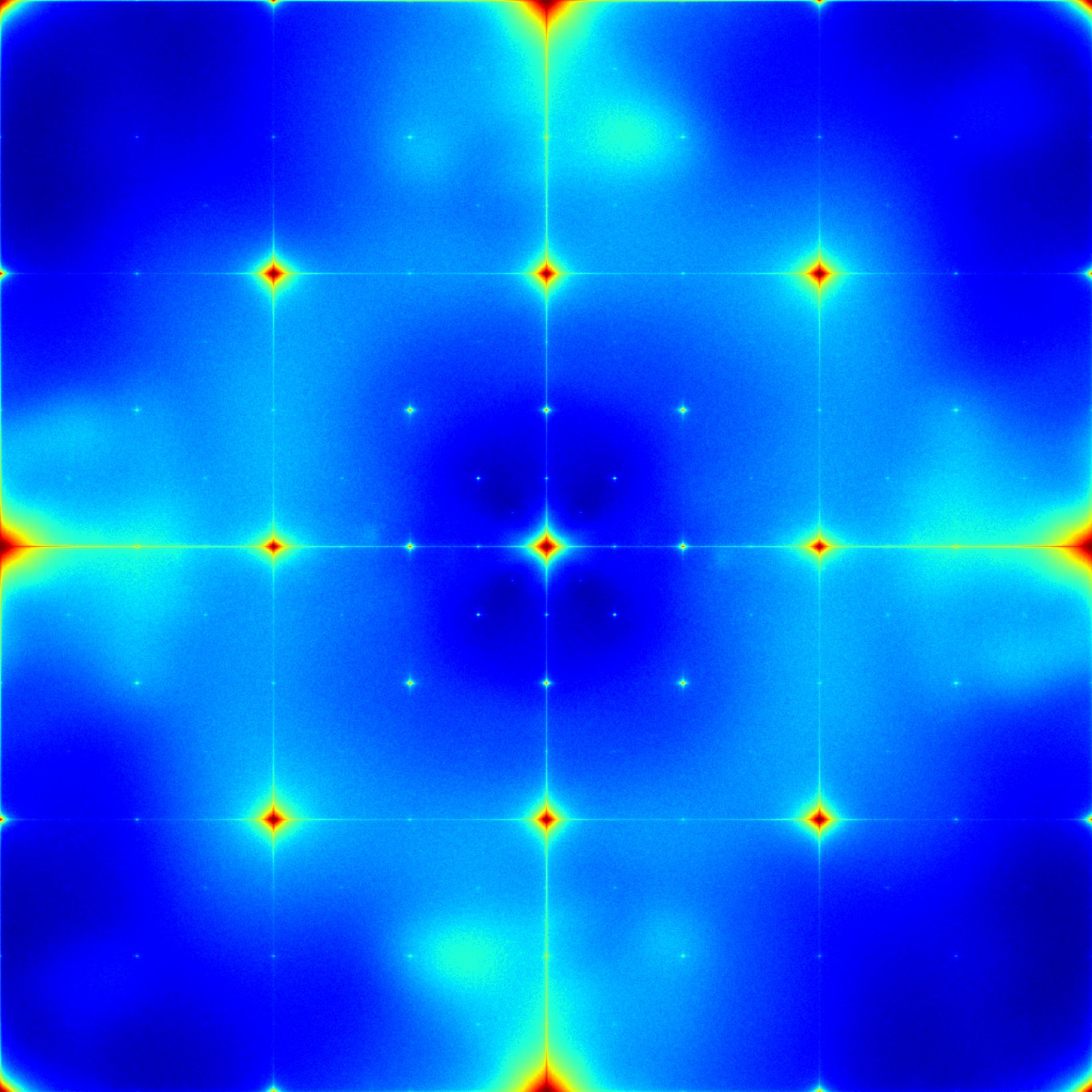}
  \caption{SWAGAN}
\end{subfigure}
\caption{Anylysis of the frequency spectra for FFHQ (pristine) and each of the generative models.}
\label{fig:frequency_artefacts}
\end{figure}

Figure~\ref{fig:frequency_artefacts} shows an analysis of the average frequency spectra for each used dataset computed over 1,000 randomly sampled images from each dataset. To provide a more clear visualization we performed a high-pass filtering over the image by first applying a Median Blur filter to the image and then subtracting it from the input image~\cite{Wang-CNNGeneratedImagesAreSurprisinglyEasyToSpotForNow-CVPR-2020}. We then perform a DFT separately on each RGB input channel (see section~\ref{sec:preprocessing}) and average the result over all randomly sampled images of the dataset. The averaged frequency spectra show some differences between the pristine input data and the data from the generative models, most notably for SWAGAN where clear periodic patterns can be observed.

\subsection{Preprocessing}
\label{sec:preprocessing}
In the preprocessing phase, the images are aligned and cropped to resolution 224$\times$224 and the DFT of each image is computed: 

\begin{description}
\item[Face Alignment]
As this study aims to detect synthetic facial images, and \eg not artefacts that might appear in the background of a synthetic image, a face is detected in each input image whereafter it is cropped out and the eyes aligned using the MTCNN~\cite{Zhang-JointFaceDetectionAndAllignmentUsingMultitaskCascadedConvolutionalNetworks-2016} algorithm. The resulting images are of resolution 224$\times$224 pixels and cropped to the face region.

\item[Frequency Domain Transformation] To transform an RGB image into the frequency spectrum we follow the approach in~\cite{Zhang-AutoGAN-WIFS-2019} and normalize the input pixels to the range [0, 1] whereafter we perform DFT for each of the 3 colour channels and obtain corresponding channels in the frequency spectrum. Thereafter, the logarithmic spectrum is computed and normalized to the range [-1, 1]. Lastly, we shift the low frequency components to the center of the spectrum. The resulting 3D array is transformed to a tensor and given as input to the classifiers. Examples of images and their corresponding spectra is given in figure~\ref{fig:images_spectra_examples}.

\begin{figure}[!htbp]
    \centering
    \includegraphics[width=\linewidth]{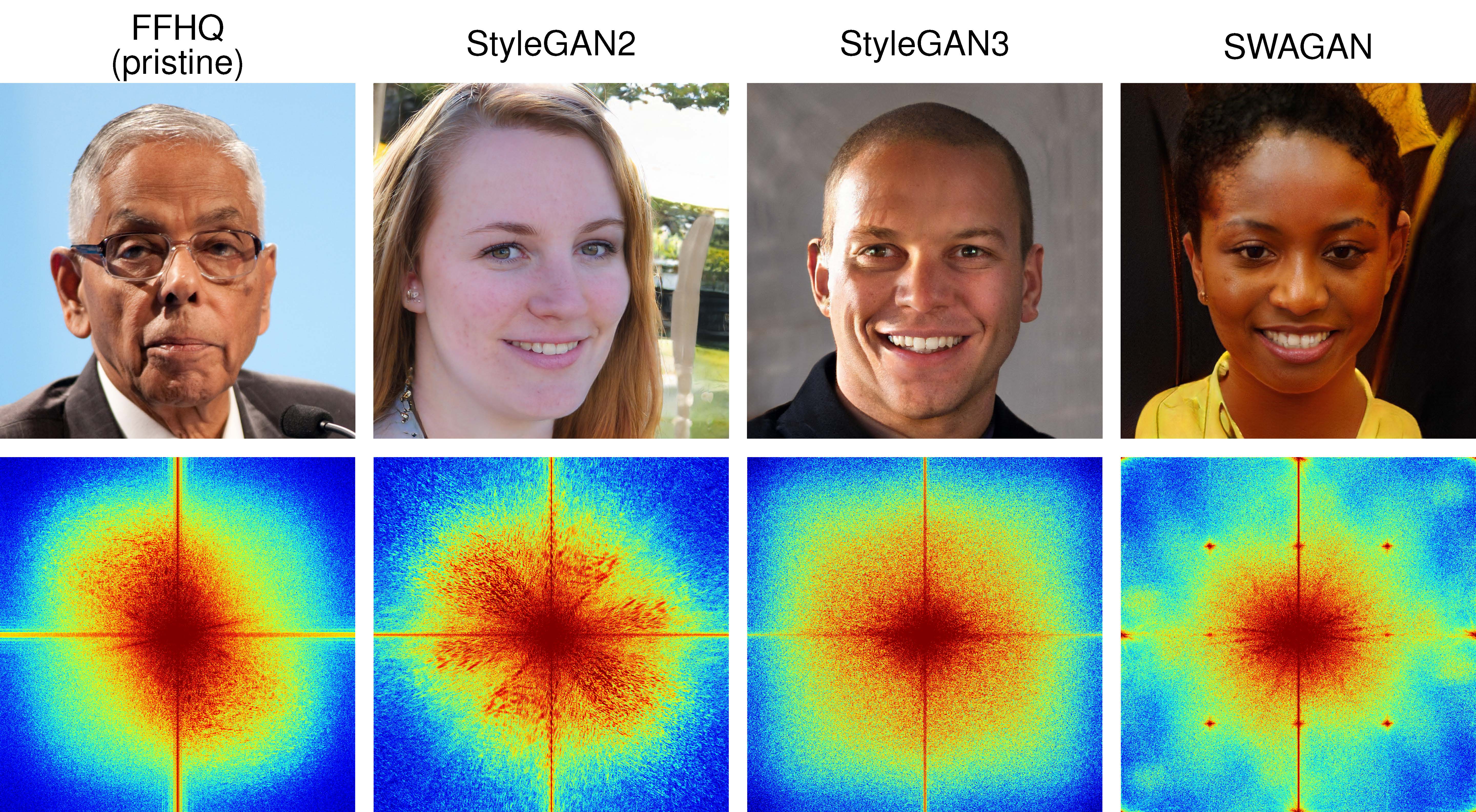}
    \caption{Examples of pristine and synthetic images and their corresponding Fourier spectrum.} \vspace{-0.4cm}
    \label{fig:images_spectra_examples}
\end{figure}

\end{description}

\subsection{Models}
In this paper the performance of the proposed architecture using CMFL is compared to several related models trained using BCE. Specifically, the following models are evaluated:

\begin{description}
\item[Dual-channel CMFL] This approach is based on the proposed architecture and CMFL loss function described in section~\ref{sec:proposed_approach}. Following~\cite{George-CMFLForRGBDFaceAntiSpoofing-CVPR-2021}, we set  $\gamma = 3$ (equation \ref{eq:cmfl}) and $\lambda=0.5$ (equation \ref{eq:aux_cmfl}).
\item[Dual-channel BCE] This approach has the same architecture as proposed in section~\ref{sec:proposed_approach}, \ie it has a separate RGB and DFT network branch, however the entire network is supervised using only the BCE loss by using the output from the joint architecture head.
\item[One-channel RGB] This approach corresponds to the single-channel equivalent of the proposed network architecture based on only an RGB input image and where the entire network is supervised by BCE.
\item[One-channel DFT] This approach is similar to the \textit{One-channel RGB} approach but operates on the computed frequency spectra of the aligned input face image instead of on the RGB image.
 \item[Fusion (RGB + DFT)] In this model, we perform a score level fusion by computing the average between the output scores of the RGB and DFT single-channel networks.
\end{description}

\subsection{Training Details}
For training, a learning rate of $1 \cdot 10^{-4}$ is used with a weight decay of  $1 \cdot 10^{-5}$. The DenseNet blocks are initialized using pre-trained weights from ImageNet and fine-tuned for 25 epochs using a batch size of 128, horizontal flips with probability 0.5, cross-validation, and the Adam optimizer~\cite{Kingma-AdamOptimizer-ICLR-2015}. The best model for each experiment on the validation set is saved and used during the evaluation. All models have been trained using NVIDIA A100 Tensor Core GPUs.

\subsection{Metrics}
To evaluate the performance of the different detection models, we show ROC curves and the following commonly used metrics:

\begin{description}
\item[ROC curve] A Receiver Operating Characteristic (ROC) curve shows the performance of a binary classification model in terms of the True Positive Rate (TPR) and False Positive Rate (FPR) at different operating points of the system. 
\item[AUC] The area under the curve (AUC) measures the area under a ROC curve and ranges from 0 to 1 where 1 indicates a perfect classifier and 0.5 represents random guessing.
\item[D-EER] Following~\cite{ISO-IEC-30107-3-PAD-metrics-2023}, the Detection Equal Error Rate (D-EER) are the rate for a binary classification task where the error rates of the two classes are equal.
\end{description}

\subsection{Experiments}
To test the different models, two protocols are proposed: 

\begin{description}
\item[Protocol I] In this protocol, the last 20,000 images of FFHQ are reserved for evaluation, whereas the remaining 35,000 and 15,000 images are reserved for training and validation, respectively. The generative models are evaluated following a leave-one-out protocol where a single model is used for training and validation in each iteration, and the remaining models are used for evaluation. In this protocol, no augmentation other than horizontal flipping at probability 0.5 is applied
\item[Protocol II] This protocol is an extension of the first protocol where Gaussian blurring and JPEG compression are applied during training. Specifically, a Gaussian blurring is applied with a $\sigma$ sampled continuously in the range [0,2], and JPEG compression is performed with a quality factor sampled discretely from: [60,70,80,90,100]. Two types of evaluations are performed; one without any augmentation and one with augmentation. During evaluation, the same augmentation is applied per image across all the experiments to make the comparisons fair.
\end{description}

The purpose of these protocols is to evaluate the models under the difficult scenario where the generative model used to generate the synthetic data is unseen during training. We focus on this scenario as it has been shown that learning to detect synthetic images from a known generative model is relatively easy and usually results in a high detection performance. 
As the purpose of this paper is to investigate the feasibility of the proposed architecture, we focus the experiments on comparing the results to related architectures trained on the same data and which utilize similar pre-processing. The code is made available to allow comparison and inclusion in benchmarks which specifically aims at comparing state-of-the-art methods for detecting synthetic face images.

\begin{figure}[!htbp]
\begin{subfigure}[t]{0.46\columnwidth}
    \centering
  \includegraphics[width=\linewidth]{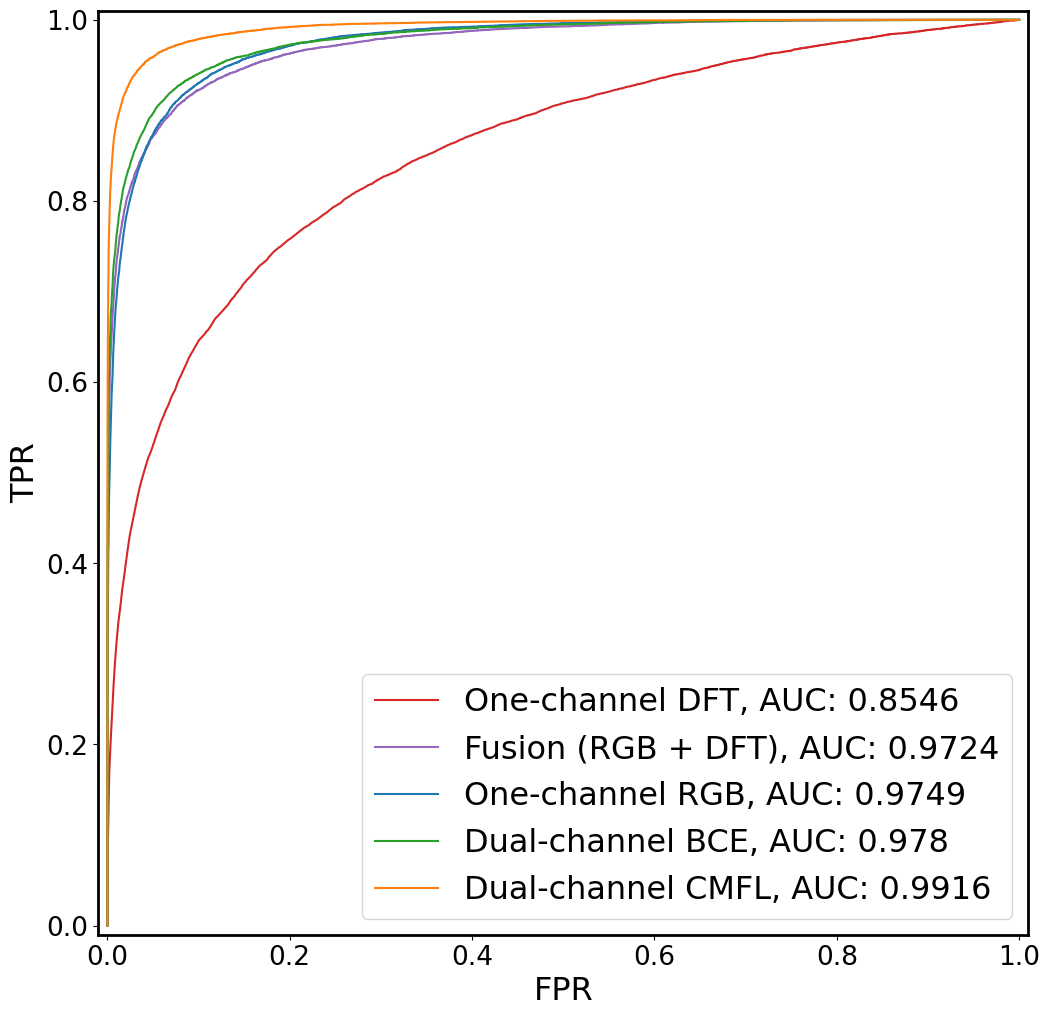}
  \caption{SWAGAN}
\end{subfigure}\quad %
\begin{subfigure}[t]{0.46\columnwidth}
    \centering
  \includegraphics[width=\linewidth]{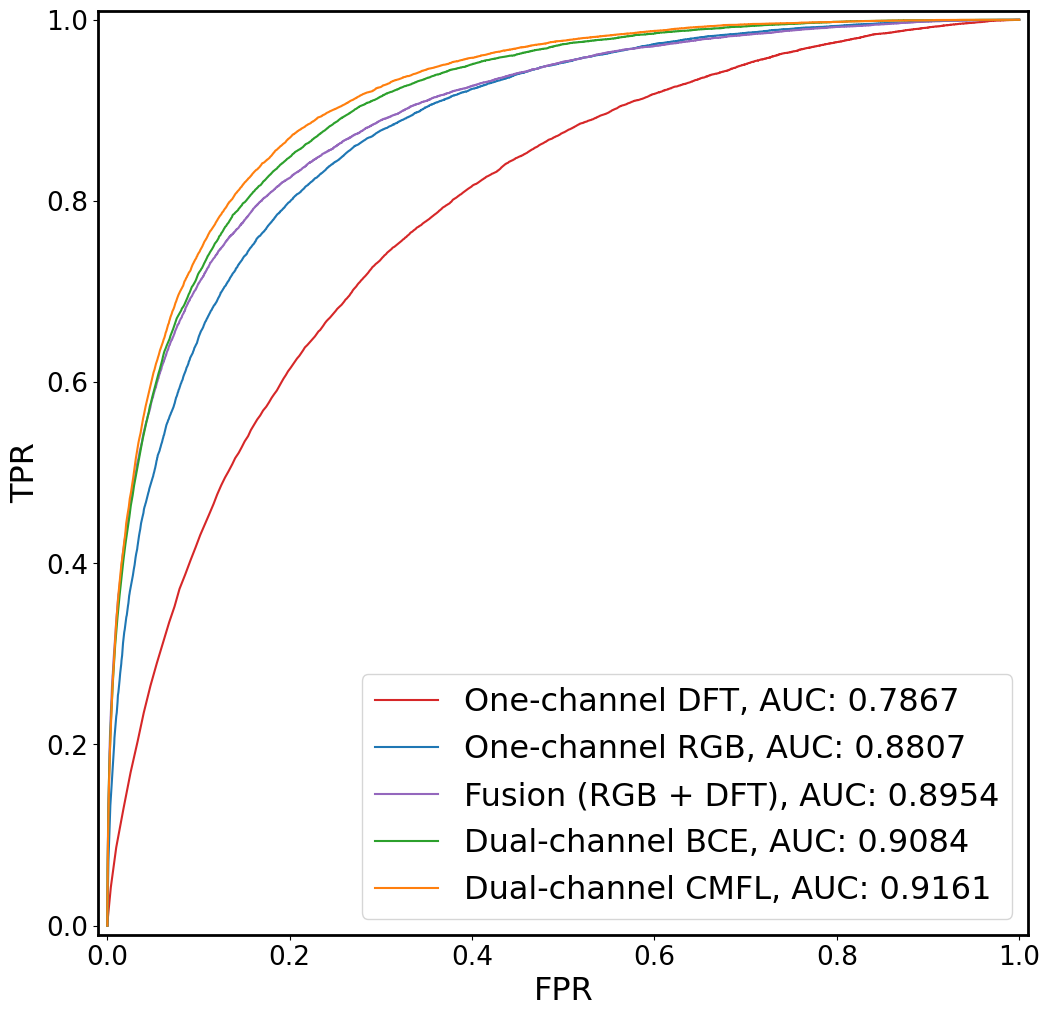}
  \caption{StyleGAN3} 
\end{subfigure}\quad %
\caption{ROC curves for protocol I when training on StyleGAN2} 
\vspace{-0.4cm}
\label{fig:stylegan2_roc_curve_protocol1}
\end{figure}

\begin{figure*}[!htbp]
\begin{subfigure}[t]{0.24\linewidth}
  \includegraphics[width=\linewidth]{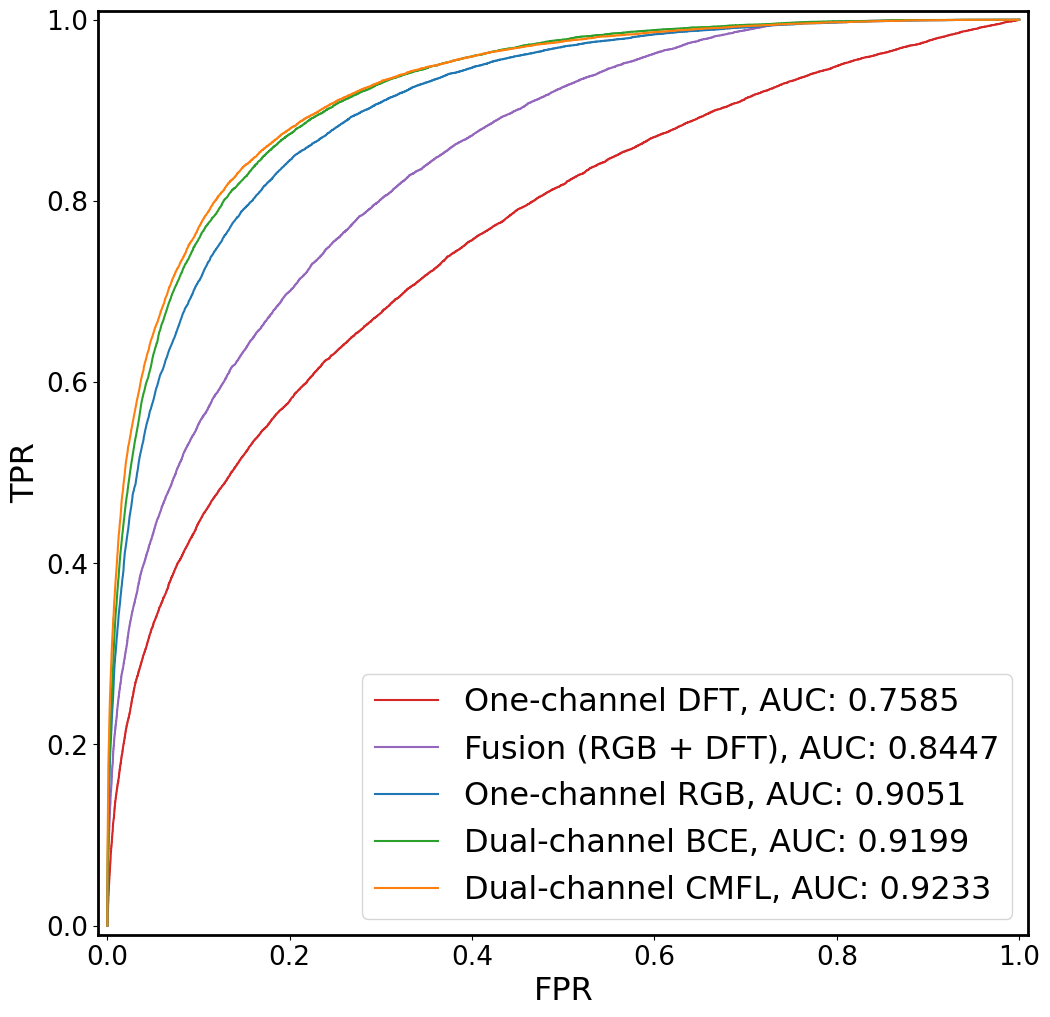}
  \caption{SWAGAN no aug.}
\end{subfigure} %
\begin{subfigure}[t]{0.24\linewidth}
  \includegraphics[width=\linewidth]{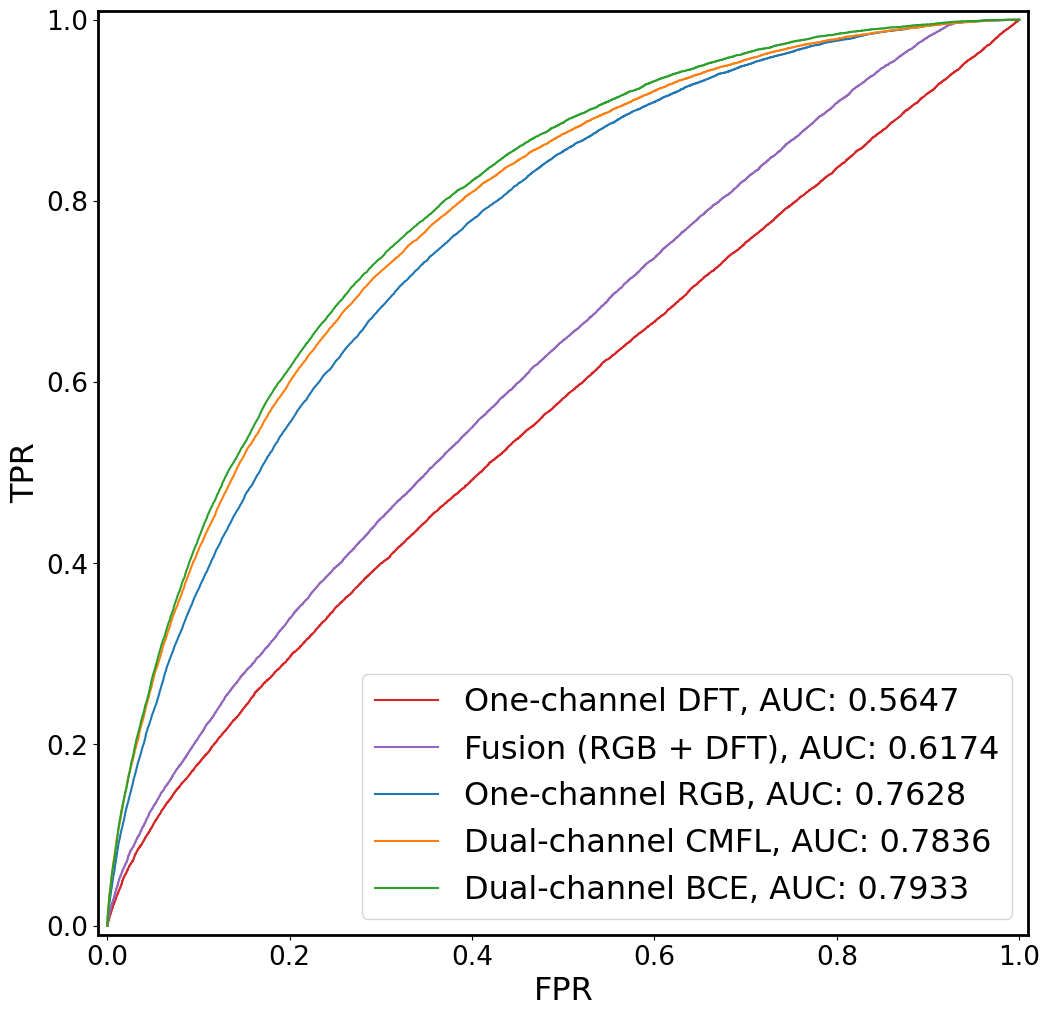}
  \caption{StyleGAN3 no aug.}
\end{subfigure} %
\centering
\begin{subfigure}[t]{0.24\linewidth}
  \includegraphics[width=\linewidth]{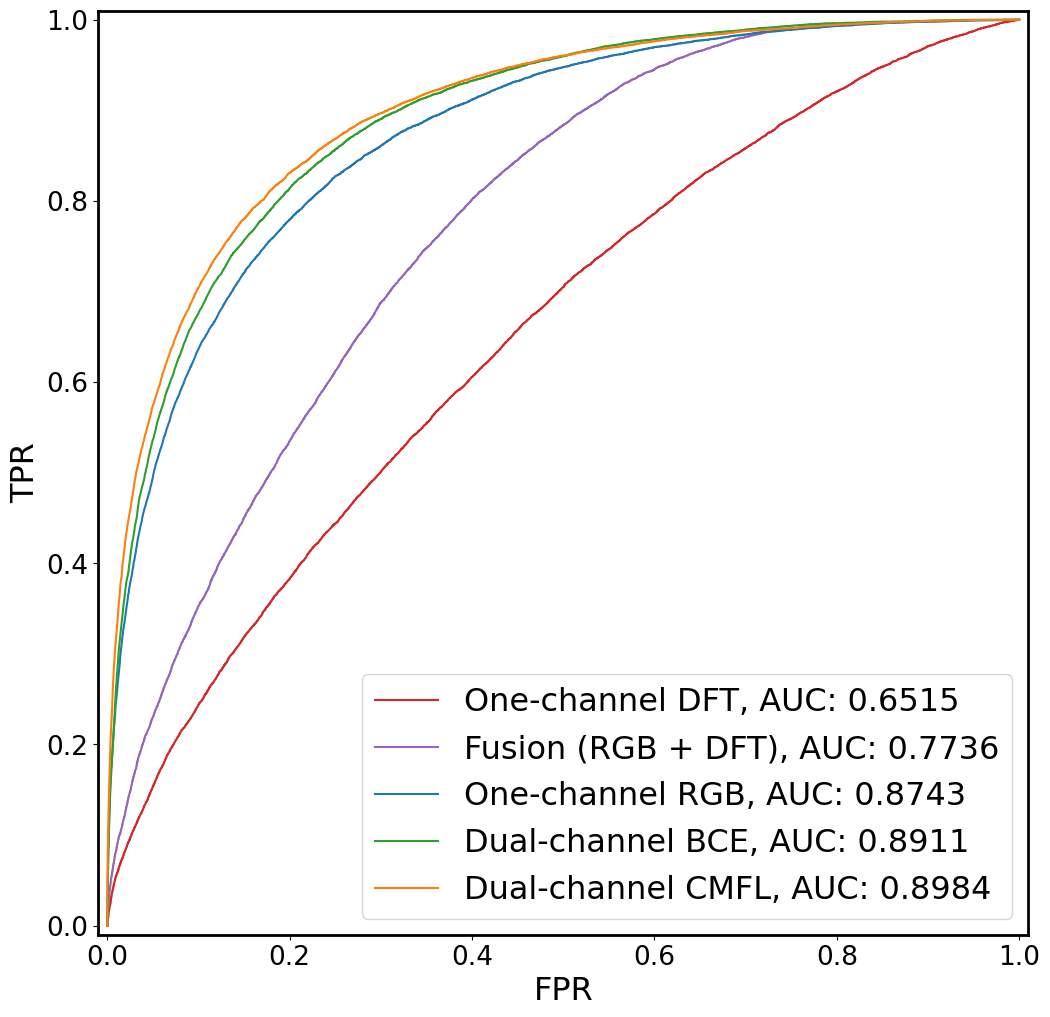}
  \caption{SWAGAN with aug.}
\end{subfigure} %
\begin{subfigure}[t]{0.24\linewidth}
  \includegraphics[width=\linewidth]{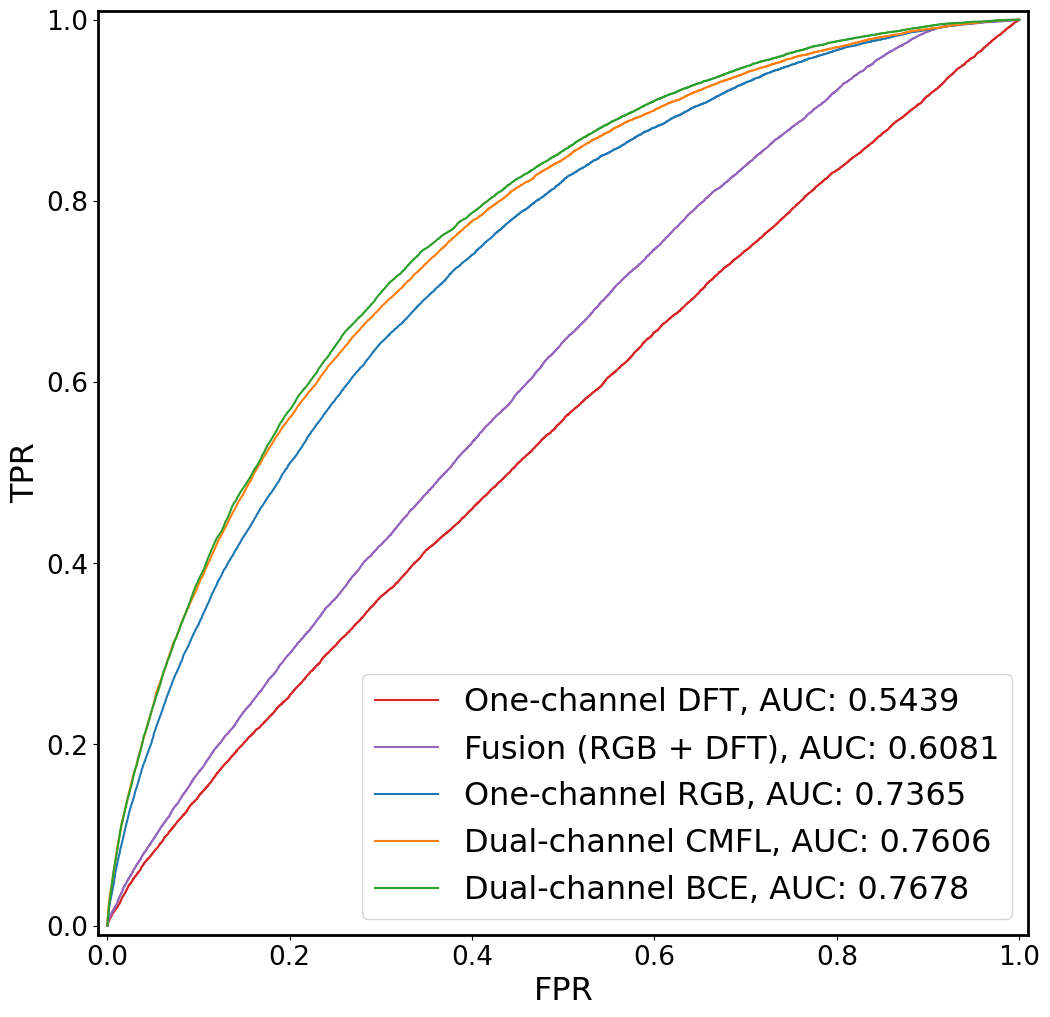}
  \caption{StyleGAN3 with aug.}
\end{subfigure} %
\caption{ROC curves for protocol II when training on StyleGAN2.} \vspace{-0.4cm}
\label{fig:stylegan2_roc_curve_protocol2}
\end{figure*} 

\begin{table*}[!htbp]
    \centering
\caption{AUC and D-EER in \% for the different models and evaluation protocols. The highlighted numbers indicate the best performance (AUC or D-EER) observed for a test protocol across all five models for a specific train and test partition.}
\begin{adjustbox}{max width=\linewidth}
   \begin{tabular}{@{\extracolsep{2pt}}llrrrrrrrrrrrr@{}} \toprule 
       & &  \multicolumn{4}{c}{\textbf{StyleGAN2}}  &
      \multicolumn{4}{c}{\textbf{StyleGAN3}} & 
      \multicolumn{4}{c}{\textbf{SWAGAN}} \\ \cmidrule{3-6} \cmidrule{7-10}  \cmidrule{11-14} 
\textbf{Model} & \textbf{Test Protocol} &   \multicolumn{2}{c}{StyleGAN3} & \multicolumn{2}{c}{SWAGAN} & \multicolumn{2}{c}{StyleGAN2} & \multicolumn{2}{c}{SWAGAN}  & \multicolumn{2}{c}{StyleGAN2} & \multicolumn{2}{c}{StyleGAN3}   \\
\textbf{} & \textbf{} &   
\multicolumn{1}{c}{AUC} & \multicolumn{1}{c}{D-EER} &   \multicolumn{1}{c}{AUC} & \multicolumn{1}{c}{D-EER} &   \multicolumn{1}{c}{AUC} & \multicolumn{1}{c}{D-EER} &   \multicolumn{1}{c}{AUC} & \multicolumn{1}{c}{D-EER} &   \multicolumn{1}{c}{AUC} & \multicolumn{1}{c}{D-EER} &   \multicolumn{1}{c}{AUC} & \multicolumn{1}{c}{D-EER} \\
\midrule
\multirow{3}{*}{One-channel RGB} & Protocol I & 0.881 & 20.04 & 0.975 & 8.29 & 0.925 & 15.42 & 0.941 & 13.41 & \textbf{0.967} & \textbf{9.40} & \textbf{0.809} & \textbf{26.74}\\
 & Protocol II (no aug.) & 0.763 & 30.84 & 0.905 & 17.79 & 0.868 & 21.53 & 0.930 & 14.90 & 0.897 & 18.57 & 0.837 & 24.24\\
 & Protocol II (with aug.) & 0.736 & 32.90 & 0.874 & 21.03 & 0.844 & 23.58 & 0.906 & 17.51 & 0.877 & 20.64 & 0.817 & 26.12\\ \midrule 
\multirow{3}{*}{One-channel DFT} & Protocol I & 0.787 & 28.29 & 0.855 & 22.40 & 0.925 & 15.32 & 0.418 & 56.19 & 0.482 & 51.79 & 0.479 & 51.95\\
 & Protocol II (no aug.) & 0.565 & 45.64 & 0.758 & 31.21 & 0.511 & 49.72 & 0.677 & 37.47 & 0.436 & 54.65 & 0.531 & 47.92\\
 & Protocol II (with aug.) & 0.544 & 47.03 & 0.652 & 39.71 & 0.531 & 48.03 & 0.586 & 43.63 & 0.522 & 48.26 & 0.523 & 48.68\\ \midrule 
\multirow{3}{*}{Fusion (RGB + DFT)} & Protocol I & 0.895 & 18.52 & 0.972 & 8.66 & 0.942 & 13.57 & 0.682 & 36.63 & 0.934 & 13.49 & 0.713 & 34.96\\
 & Protocol II (no aug.) & 0.617 & 42.51 & 0.845 & 24.65 & 0.601 & 44.29 & 0.793 & 29.61 & 0.569 & 46.22 & 0.605 & 43.32\\
 & Protocol II (with aug.) & 0.608 & 43.21 & 0.774 & 30.57 & 0.622 & 42.49 & 0.725 & 34.61 & 0.599 & 43.81 & 0.575 & 45.57\\ \midrule 
\multirow{3}{*}{Dual-channel BCE} & Protocol I & 0.908 & 17.48 & 0.978 & 7.51 & 0.982 & 6.76 & \textbf{0.951} & \textbf{12.01} & 0.953 & 11.21 & 0.722 & 33.75\\
 & Protocol II (no aug.) & \textbf{0.793} & \textbf{28.10} & 0.920 & 16.14 & 0.883 & 19.90 & 0.942 & 13.11 & 0.900 & 18.44 & 0.858 & 22.61\\
 & Protocol II (with aug.) & \textbf{0.768} & \textbf{30.04} & 0.891 & 19.34 & 0.854 & 22.66 & 0.917 & 16.33 & 0.873 & 21.02 & 0.832 & 24.87\\ \midrule 
\multirow{3}{*}{Dual-channel CMFL} & Protocol I & \textbf{0.916} & \textbf{16.47} & \textbf{0.992} & \textbf{4.45} & \textbf{0.984} & \textbf{6.38} & 0.899 & 18.28 & 0.952 & 11.59 & 0.693 & 35.87\\
 & Protocol II (no aug.) & 0.784 & 28.86 & \textbf{0.923} & \textbf{15.65} & \textbf{0.894} & \textbf{18.96} & \textbf{0.947} & \textbf{12.60} & \textbf{0.909} & \textbf{17.23} & \textbf{0.864} & \textbf{21.85}\\
 & Protocol II (with aug.) & 0.761 & 30.88 & \textbf{0.898} & \textbf{18.38} & \textbf{0.867} & \textbf{21.61} & \textbf{0.923} & \textbf{15.75} & \textbf{0.882} & \textbf{20.22} & \textbf{0.838} & \textbf{24.21}\\ \bottomrule 
    \end{tabular}
\label{tab:performance_all_models}
\end{adjustbox}{}
\end{table*}

\section{Results}
\label{sec:results}
In this section, we analyze the results for the two training and evaluation protocols. Table~\ref{tab:performance_all_models} shows the AUC and D-EER for all protocols and models. Furthermore, we highlight the results for protocol I and protocol II when training on FFHQ and StyleGAN2 using ROC curves (see figure~\ref{fig:stylegan2_roc_curve_protocol1} and~\ref{fig:stylegan2_roc_curve_protocol2}).

As seen in table~\ref{tab:performance_all_models} the dual-channel approaches performs best across most test protocols and training scenarios. Furthermore, and as expected, it is easier to detect images across all models when the images are not augmented or blurred during evaluation. Hence for protocol I we observe that the best model achieves AUC $>$ 0.90 across almost all experiments with the exception being when trained on SWAGAN and evaluated on StyleGAN3 where the best model achieves a AUC of $\approx 0.81$. Looking at the ROC-curve for protocol I in figure~\ref{fig:stylegan2_roc_curve_protocol1} when training on StyleGAN2 and evaluating on StyleGAN3 and SWAGAN, we can see that the proposed architecture with CMFL is the best model. For protocol II, shown in figure \ref{fig:stylegan2_roc_curve_protocol2}, we can observe that the proposed architecture using CMFL is better when evaluated on SWAGAN but that the proposed architecture with BCE is slightly better when evaluated on StyleGAN3. Looking accross all the experiments, we can see that using BCE and CMFL, in general, is quite similar but that using CMFL is better in most cases. Another finding is that DFT is not always capable of generalizing to unseen models. This is, especially, notable for protocol II when training and evaluating with realistic post-processing operations. Interestingly, poor performance for the one-channel DFT model is, especially, observed when training on SWAGAN and evaluating on StyleGAN2 and StyleGAN3 where the performance is close to random. This performance is most likely due to the SWAGAN images being easy to detect in the frequency domain due to clear artefacts (see figure~\ref{fig:frequency_artefacts}) which are not as heavily present in the images generated by the StyleGAN models. Despite this, we can observe that even in cases where the DFT approach performs poor, the dual-channel approaches are still capable of achieving good performance due to the RGB channel which in general is capable of achieving high detection performance. Nevertheless, we can observe that the dual-channel approaches achieve superior performance to both the single-channel DFT and RGB approaches. Additionally, the results show that it is better to train the two channels jointly rather than doing a simple score fusion between the two single-channel models. Furthermore, we can observe that in cases where both the RGB and DFT single-channel approaches have good performance, \ie both have an AUC $>$ 0.70, supervising the network using CMFL achieves the best performance. This indicates that CMFL can generalize better in cases where discriminative information is present in both domains. The results shown in this section are limited to GAN-based images and to models trained and evaluated with FFHQ data.

\section{Conclusion}
\label{sec:conclusion}
In this work, we proposed a two-channel neural network architecture for detecting synthetic face images with one channel using information from the visible spectrum (RGB) and the other from the frequency spectrum computed separately for each colour channel using Discrete Fourier Transform. Furthermore, we proposed to supervise the network using Cross Modal Focal Loss which is capable of better modulating individual channel contribution. We evaluated the architecture on three popular generative models and compared the performance to single-channel architectures and to the same architecture supervised fully using Binary Cross Entropy. Through cross-model experiments we showed that the proposed architecture using Cross Modal Focal Loss, generally, outperformed the other models, especially in cases where images were detectable in both the visible and frequency spectra. Our studies emphasize the need to continue to understand and explore how novel loss functions can be integrated and used to detect synthetic face images and to explore the performance of detection models in challenging and realistic scenarios. 

\section{Acknowledgement}
This research work has been partially funded by the German Federal Ministry of Education and Research and the Hessian Ministry of Higher Education, Research, Science and the Arts within their joint support of the National Research Center for Applied Cybersecurity ATHENE and the European Union's Horizon 2020 research and innovation programme under the Marie Sk\l{}odowska-Curie grant agreement No. 860813 - TReSPAsS-ETN. Additionally, the authors would like to thank Anjith George for helpful discussions. 

{\small
\bibliographystyle{ieee}
\bibliography{egbib}

\begin{thebibliography}{10}\itemsep=-1pt

\bibitem{ffhq}
{Flickr-Faces-HQ Dataset (FFHQ)}.
\newblock \url{https://github.com/NVlabs/ffhq-dataset}.
\newblock Last accessed: 2024-01-23.

\bibitem{Chen-JoiuntSpatialFrequencyDomainNetwork-arxiv-2020}
Z.~Chen and H.~Yang.
\newblock Manipulated face detector: Joint spatial and frequency domain attention network.
\newblock {\em arXiv e-prints}, page arXiv:2005.02958v1, 05 2020.

\bibitem{Frank-FrequencyAnalysisForDeepFakeImageRecognition-ICML-2020}
J.~Frank, T.~Eisenhofer, L.~Sch\"{o}nherr, A.~Fischer, et~al.
\newblock Leveraging frequency analysis for deep fake image recognition.
\newblock In {\em 37th Int'l Conf. on Machine Learning ({ICML})}, 2020.

\bibitem{Fu-RobustDualChannelGANDetectionBasedOnFilters-CISP-2019}
Y.~Fu, T.~Sun, X.~Jiang, K.~Xu, and P.~He.
\newblock Robust {GAN}-face detection based on dual-channel {CNN} network.
\newblock In {\em Int'l Congress on Image and Signal Processing, BioMedical Engineering and Informatics ({CISP-BMEI})}, pages 1--5, 2019.

\bibitem{Gal-Swagan-ACM-2021}
R.~Gal, D.~C. Hochberg, A.~Bermano, and D.~Cohen-Or.
\newblock {SWAGAN}: A style-based wavelet-driven generative model.
\newblock {\em ACM Trans. Graph.}, 40(4), jul 2021.

\bibitem{George-CMFLForRGBDFaceAntiSpoofing-CVPR-2021}
A.~George and S.~Marcel.
\newblock Cross modal focal loss for {RGBD} face anti-spoofing.
\newblock In {\em Conf. on Computer Vision and Pattern Recognition (CVPR)}, pages 7878--7887, 2021.

\bibitem{Gragnaniello-DetectionOfAIGeneratedFaces-Springer-2022}
D.~Gragnaniello, F.~Marra, and L.~Verdoliva.
\newblock Detection of {AI}-generated synthetic faces.
\newblock In {\em Handbook of Digital Face Manipulation and Detection: From DeepFakes to Morphing Attacks}, Advances in Computer Vision and Pattern Recognition, pages 191--212. Springer Verlag, 2022.

\bibitem{Hsu-DeepfakeAlgorithmUsingMultiplNoiseModalitiesWithTwoBranchPrediction-APSIPAASC}
H.-W. Hsu and J.-J. Ding.
\newblock Deepfake algorithm using multiple noise modalities with two-branch prediction network.
\newblock In {\em Asia-Pacific Signal and Information Processing Association Annual Summit and Conference ({APSIPA ASC})}, pages 1662--1669, 2021.

\bibitem{Huang-DenseNet-CVPR-2017}
G.~Huang, Z.~Liu, L.~V.~D. Maaten, and K.~Q. Weinberger.
\newblock Densely connected convolutional networks.
\newblock In {\em Conf. on Computer Vision and Pattern Recognition ({CVPR})}, pages 2261--2269, 2017.

\bibitem{ISO-IEC-30107-3-PAD-metrics-2023}
{ISO/IEC JTC1 SC37 Biometrics}.
\newblock {\em {ISO/IEC} 30107-3. Information Technology - Biometric presentation attack detection - Part 3: Testing and Reporting}.
\newblock International Organization for Standardization, 2023.

\bibitem{Joshi-SyntheticDataInHumanAnalysisSurvey-2022}
I.~Joshi, M.~Grimmer, C.~Rathgeb, C.~Busch, F.~Bremond, and A.~Dantcheva.
\newblock Synthetic data in human analysis: A survey.
\newblock {\em IEEE Trans. on Pattern Analysis and Machine Intelligence}, pages 1--20, 2024.

\bibitem{Kabbani-EGAIN-2022-EUVIP}
W.~Kabbani, M.~Grimmer, and C.~Busch.
\newblock {EGAIN}: {Extended} {GAn} {INversion}.
\newblock In {\em 10th European Workshop on Visual Information Processing ({EUVIP})}, pages 1--6, 2022.

\bibitem{Karras-StyleGAN3-NIPS-2021}
T.~Karras, M.~Aittala, S.~Laine, E.~H\"{a}rk\"{o}nen, et~al.
\newblock Alias-free generative adversarial networks.
\newblock In {\em Advances in Neural Information Processing Systems}, volume~34, pages 852--863, 2021.

\bibitem{Karras-StyleGAN2-CVPR-2020}
T.~Karras, S.~Laine, M.~Aittala, J.~Hellsten, et~al.
\newblock Analyzing and improving the image quality of {StyleGAN}.
\newblock In {\em Conf. on Computer Vision and Pattern Recognition ({CVPR})}, pages 8107--8116, 2020.

\bibitem{Kietzmann-DeepfakesTrickOrTreat-BusinessHorizons-2020}
J.~Kietzmann, L.~W. Lee, I.~P. McCarthy, and T.~C. Kietzmann.
\newblock Deepfakes: Trick or treat?
\newblock {\em Business Horizons}, 63(2):135--146, 2020.

\bibitem{Kingma-AdamOptimizer-ICLR-2015}
D.~P. Kingma and J.~Ba.
\newblock Adam: A method for stochastic optimization.
\newblock In {\em Int'l Conf. on Learning Representations ({ICLR})}, 2015.

\bibitem{Lin-FocalLoss-ICCV-2017}
T.-Y. Lin, P.~Goyal, R.~Girshick, K.~He, and P.~Doll\'ar.
\newblock Focal loss for dense object detection.
\newblock In {\em Int'l Conf. on Computer Vision ({ICCV})}, pages 2999--3007, 2017.

\bibitem{Masi-TwoBranchRecurrentNetworkForIsolatingDeepfakesInVideos-ECCV-2020}
I.~Masi, A.~Killekar, R.~M. Mascarenhas, S.~P. Gurudatt, and W.~AbdAlmageed.
\newblock Two-branch recurrent network for isolating deepfakes in videos.
\newblock In {\em European Conf. on Computer Vision {(ECCV)}}, pages 667--684, 2020.

\bibitem{Matern-ExploitingVisualArtefacts-WACVW-2019}
F.~Matern, C.~Riess, and M.~Stamminger.
\newblock Exploiting visual artifacts to expose deepfakes and face manipulations.
\newblock In {\em {IEEE} Winter Applications of Computer Vision Workshops ({WACVW})}, pages 83--92, 2019.

\bibitem{Nightingale-HumanDetectionOfAiSynthesizedImages-PNAS-2022}
S.~J. Nightingale and H.~Farid.
\newblock {AI}-synthesized faces are indistinguishable from real faces and more trustworthy.
\newblock {\em Proceedings of the National Academy of Sciences}, 119(8):e2120481119, 2022.

\bibitem{Rathgeb2019-beaty}
C.~Rathgeb, A.~Dantcheva, and C.~Busch.
\newblock Impact and detection of facial beautification in face recognition: An overview.
\newblock {\em {IEEE} Access}, 2019.

\bibitem{roessler2019faceforensics}
A.~R\"ossler, D.~Cozzolino, L.~Verdoliva, C.~Riess, et~al.
\newblock Faceforensics++: Learning to detect manipulated facial images.
\newblock In {\em Intl. Conf. of Computer Vision (ICCV'19)}, 2019.

\bibitem{Wang-CNNGeneratedImagesAreSurprisinglyEasyToSpotForNow-CVPR-2020}
S.-Y. Wang, O.~Wang, R.~Zhang, A.~Owens, and A.~A. Efros.
\newblock {CNN}-generated images are surprisingly easy to spot… for now.
\newblock In {\em Conf. on Computer Vision and Pattern Recognition ({CVPR})}, pages 8692--8701, 2020.

\bibitem{Yang-ExposingGANSynthesizedFacesUsingLandmarks-ACM-2019}
X.~Yang, Y.~Li, H.~Qi, and S.~Lyu.
\newblock Exposing {GAN}-synthesized faces using landmark locations.
\newblock In {\em {ACM} Workshop on Information Hiding and Multimedia Security}, pages 113--–118, 2019.

\bibitem{Zhang-JointFaceDetectionAndAllignmentUsingMultitaskCascadedConvolutionalNetworks-2016}
K.~Zhang, Z.~Zhang, Z.~Li, and Y.~Qiao.
\newblock {Joint Face Detection and Alignment Using Multitask Cascaded Convolutional Networks}.
\newblock {\em {IEEE} Signal Processing Letters}, 23(10):1499--1503, 2016.

\bibitem{Zhang-AutoGAN-WIFS-2019}
X.~Zhang, S.~Karaman, and S.-F. Chang.
\newblock Detecting and simulating artifacts in {GAN} fake images.
\newblock In {\em {IEEE} Int'l Workshop on Information Forensics and Security ({WIFS})}, pages 1--6, 2019.

\bibitem{Zhou-TwoStreamTamperedFaceDetection-CVPRW-2017}
P.~Zhou, X.~Han, V.~I. Morariu., and L.~S. Davis.
\newblock Two-stream neural networks for tampered face detection.
\newblock In {\em Conf. on Computer Vision and Pattern Recognition Workshops ({CVPRW})}, pages 1831--1839, 2017.

\end{thebibliography}
}
\end{document}